\documentclass{article}
\usepackage{PRIMEarxiv}

\usepackage{amsmath,amsfonts,bm}

\def\eqref#1{equation~\ref{#1}}

\def\1{\bm{1}}

\DeclareMathAlphabet{\mathsfit}{\encodingdefault}{\sfdefault}{m}{sl}
\SetMathAlphabet{\mathsfit}{bold}{\encodingdefault}{\sfdefault}{bx}{n}

\usepackage[utf8]{inputenc} 
\usepackage[T1]{fontenc}    
\usepackage{hyperref}       
\usepackage{url}            
\usepackage{booktabs}       
\usepackage{amsfonts}       
\usepackage{nicefrac}       
\usepackage{microtype}      
\usepackage{lipsum}
\usepackage{graphicx}
\graphicspath{{media/}}     
\usepackage{colortbl}
\usepackage{microtype}
\usepackage{graphicx}
\usepackage{multirow}
\usepackage[inline, shortlabels]{enumitem}
\usepackage{amssymb}
\usepackage{mathtools}
\usepackage{amsthm}
\usepackage{subfigure}

\theoremstyle{definition}

\theoremstyle{remark}

\newcommand{\ie}{\textit{i}.\textit{e}., }
\newcommand{\eg}{\textit{e}.\textit{g}., }

\usepackage{algorithm, algorithmic, bm}
\usepackage{amsthm,amsmath,amssymb}
\usepackage{pifont}

\usepackage{wrapfig,lipsum,booktabs}

\title{Task-Aware Low-Rank Adaptation of Segment Anything Model
}

\author{Xuehao Wang$^{1}$, 
Feiyang Ye$^{1,2}$,
Yu Zhang$^{1,}$\thanks{Corresponding author}\\
  $^1$ Southern University of Science and Technology \\
  $^2$ University of Technology Sydney \\
  \texttt{\{xuehaowangfi,feiyang.ye.uts,yu.zhang.ust\}@gmail.com}
}

\begin{document}
\maketitle
\begin{abstract}
The Segment Anything Model (SAM), with its remarkable zero-shot capability, has been proven to be a powerful foundation model for image segmentation tasks, which is an important task in computer vision.
However, the transfer of its rich semantic information to multiple different downstream tasks remains unexplored.
In this paper, we propose the Task-Aware Low-Rank Adaptation (TA-LoRA) method, which enables SAM to work as a foundation model for multi-task learning.
Specifically, TA-LoRA injects an update parameter tensor into each layer of the encoder in SAM and leverages a low-rank tensor decomposition method to incorporate both task-shared and task-specific information. 
Furthermore, we introduce modified SAM (mSAM) for multi-task learning where we remove the prompt encoder of SAM and use task-specific no mask embeddings and mask decoder for each task.
Extensive experiments conducted on benchmark datasets substantiate the efficacy of TA-LoRA in enhancing the performance of mSAM across multiple downstream tasks.
\end{abstract}

\section{Introduction}
\label{sec:intro}

Empowered by large-scale datasets and computational advancements, large foundation models have revolutionized natural language processing and multi-modal learning, exhibiting remarkable zero-shot capabilities~\cite{kenton2019bert,lewis2020bart,radford2018improving,radford2019language,brown2020language,radford2021learning}.
Recently, the Segment Anything Model (SAM)~\cite{kirillov2023segany}, a foundation model in computer vision for image segmentation, achieves exceptional zero-shot performance across diverse tasks through training on a large-scale dataset of $11$ million samples. Efforts have been dedicated to expanding the zero-shot capability of SAM to various tasks, including high-quality segmentation~\cite{sam_hq},
object tracking~\cite{yang2023track}, 
medical image processing~\cite{ma2024segment,huang2024segment},
personalize segmentation~\cite{zhang2023personalize}, and
remote sensing~\cite{shankar2023semantic}.

Though SAM has achieved remarkable performance in diverse segmentation tasks in previous studies, those studies only consider the SAM as a foundation model for one single task, \ie, cutting-edge image segmentation. In many real-world computer vision applications, there is usually more than one task to be considered simultaneously, such as depth estimation and surface normal estimation tasks in dense scene understanding. Therefore, the ability of SAM as a foundation model for multiple computer vision tasks has not been fully explored. Although tasks such as depth estimation, and surface normal estimation, are different from segmentation tasks, previous work on multi-task learning \cite{misra2016cross,liu2019end,invpt2022,taskprompter2023,zamir2018taskonomy,liu2022polyhistor} has shown that these tasks are relevant and can even share knowledge and benefit each other during the training process. This motivates us to adopt SAM as a powerful foundation model for multi-tasking learning problems in computer vision.

To adapt large foundation models such as SAM to downstream tasks, a standard approach is to do fine-tuning. Various parameter-efficient fine-tuning (PEFT) methods have been proposed to achieve comparable performance with full fine-tuning while using only a few trainable parameters~\cite{houlsby2019parameter,lin2020exploring,li2021prefix,lester2021power,zhong2023convolution,kopiczko2023elora,wu2023mole,hu2021lora,ding2023sparse,valipour2023dylora}. Among them, the Low-Rank Adaptation (LoRA) method~\cite{hu2022lora} has received extensive attentions due to its parameter efficiency and favorable performance.
However, since the LoRA method is designed for single-task settings, it ignores the task-shared or task-specific information when we need to fine-tune a foundation model for multiple tasks simultaneously. This limitation hampers its effectiveness in multi-task learning scenarios, leading to suboptimal outcomes. Moreover, fine-tuning a separate LoRA for each task results in a linear increase in terms of the number of trainable parameters with the number of tasks, revealing a lack of parameter efficiency.

\begin{figure}[tb]
  \centering
  \subfigure[Outputs of Original SAM.]
{\label{fig:source}\includegraphics[width=0.85\textwidth]{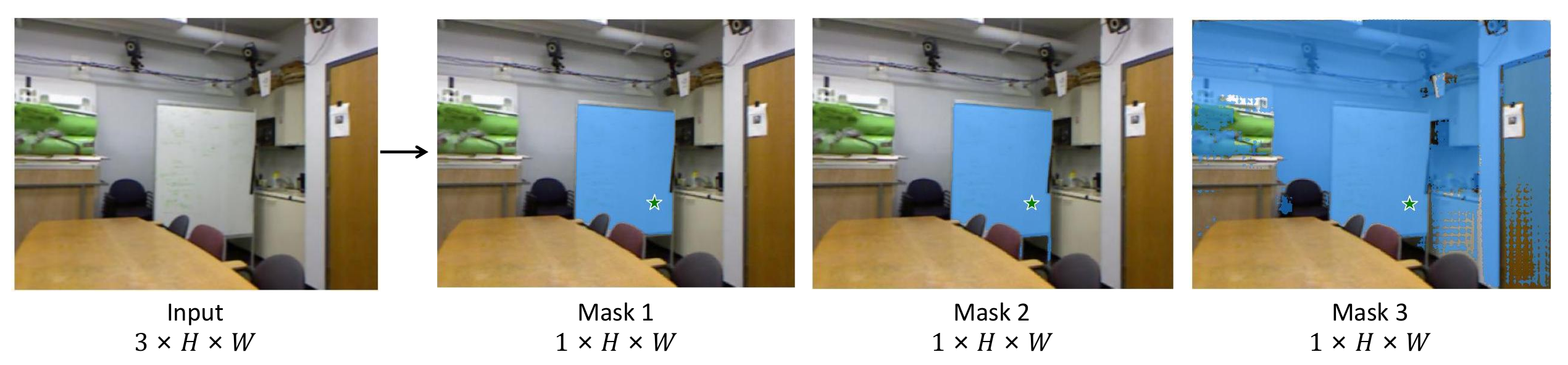}}
\subfigure[Outputs of mSAM with TA-LoRA for multiple tasks.]
{\label{fig:target}\includegraphics[width=0.85\textwidth]{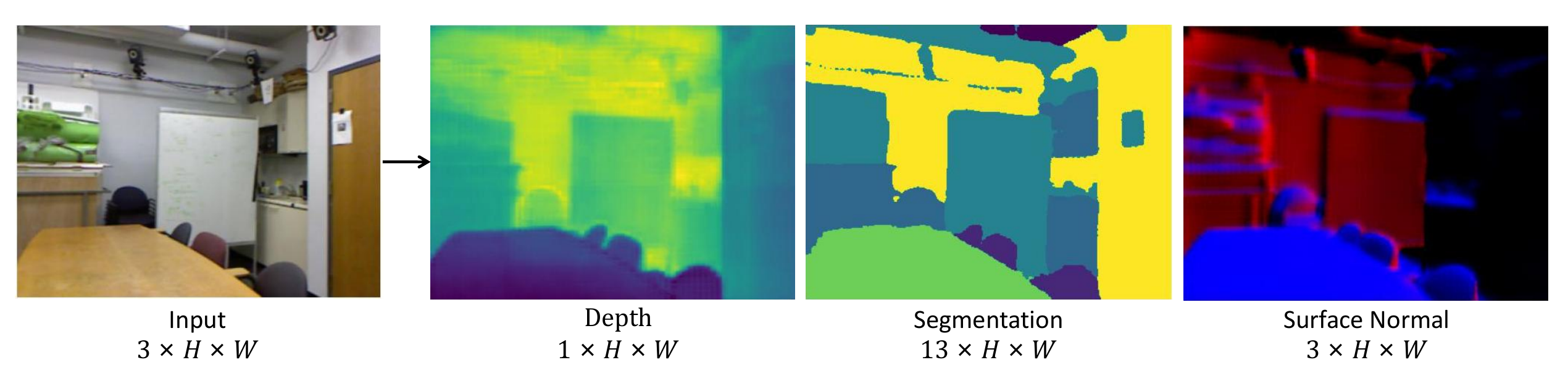}}
  \caption{Comparison between the outputs of (a) the original SAM and (b) the mSAM proposed in this paper. It can be observed that the original SAM outputs segmentation results at three different levels with an identical number of channels. In contrast, the mSAM with TA-LoRA can produce results for different tasks with varying numbers of channels.
  }
  
    \vskip -0.1in
  \label{fig:change_of_target}
\end{figure}

To address the above limitations, we propose a novel PEFT method named Task-Aware Low-Rank Adaption (TA-LoRA) for applying SAM as a foundation model for multi-tasking learning. Specifically, the proposed TA-LoRA method injects an update parameter tensor into each layer of the encoder in SAM, where each slice of the update parameter tensor acts as an update parameter matrix for the corresponding task.
By making a low-rank assumption on each update parameter tensor, we apply a low-rank tensor decomposition to it to capture both task-shared and task-specific information. For the proposed TA-LoRA method, the number of learnable parameters in the update parameter tensor exhibits a sublinear growth with respect to the number of tasks, showcasing exceptional parameter efficiency in contrast to the linear growth in the original LoRA method when directly applied to multiple tasks. 
As shown in Figure \ref{fig:change_of_target}, to enable SAM to adapt to downstream tasks with varying properties and retain the pre-training information, the proposed TA-LoRA method freezes the parameters of the heavyweight encoder and only fine-tune update parameter tensors.

The main contributions of this paper are four-fold.
\begin{itemize}

\item[$\bullet$] We propose TA-LoRA, a novel parameter-efficient fine-tuning method, which employs low-rank decomposition on the update parameter tensor to enhance model performance by effectively learning both task-shared and task-specific information simultaneously.

\item[$\bullet$] The proposed TA-LoRA method exhibits sublinear growth in the required trainable parameters with the increase in the number of tasks, demonstrating superior parameter efficiency.

\item[$\bullet$] We propose to employ different numbers of no mask embeddings to enable the adaptation of SAM to downstream tasks with different numbers of channels.

\item[$\bullet$] Extensive experiments on benchmark datasets show the exceptional performance of the proposed TA-LoRA method.
\end{itemize}

\section{Related Works}

\paragraph{Pretrained Foundation Models.}
The pre-training of large-scale foundation models has attracted considerable attention due to their powerful zero-shot capability and generalization ability, particularly in the domains of natural language processing (NLP) and multi-modal learning. Representative foundation models include 
BERT~\cite{kenton2019bert}, BART~\cite{lewis2020bart}, the GPT series~\cite{radford2018improving,radford2019language,brown2020language}, and CLIP~\cite{radford2021learning}.
Inspired by those models, the SAM~\cite{kirillov2023segany} has devised an advanced data engine, which collected an extensive dataset comprising 11 million image masks, followed by training a segmentation foundation model. This impressive foundation model demonstrates the ability to accurately segment images by utilizing diverse geometric prompts and exhibits remarkable zero-shot generalization across various downstream image segmentation tasks. Several works are proposed to accelerate SAM. For example, Fast-SAM~\cite{zhao2023fast} introduces a CNN-based architecture to enjoy faster inference speed, while MobileSAM~\cite{zhang2023faster} introduces a mobile-friendly version of SAM by substituting the heavyweight image encoder with a lightweight one.

\paragraph{Application of Segment Anything Model.}
The remarkable zero-shot generalization ability exhibited by SAM showcases its immense potential for both research and industrial applications. This potential has captured the attention of researchers, leading to numerous attempts to explore and harness its capabilities for various downstream tasks, including
high-quality segmentation~\cite{sam_hq},
object tracking~\cite{yang2023track}, 
medical image processing~\cite{ma2024segment,huang2024segment},
personalize segmentation~\cite{zhang2023personalize}, and
remote sensing~\cite{shankar2023semantic}.
In contrast to those modifications targeted at a single downstream task, the proposed TA-LoRA for SAM aims to simultaneously learn multiple downstream tasks and extract shared knowledge from these tasks to enhance the performance of the mSAM.

\paragraph{Parameter-Efficient Fine-tuning.}
To address the parameter and computational efficiency concerns during fine-tuning of large-scale pre-trained foundation models, various PEFT methods have been proposed,
including adapter-based methods~\cite{houlsby2019parameter,lin2020exploring}, prompt tuning methods~\cite{li2021prefix,lester2021power}, and LoRA-based methods~\cite{zhong2023convolution,kopiczko2023elora,wu2023mole,hu2021lora,ding2023sparse,valipour2023dylora}. 
Specially, LoRA~\cite{hu2021lora} introduces trainable low-rank matrices into transformer layers to approximate update parameter matrix, Conv-LoRA~\cite{zhong2023convolution} inserts Mixture of Experts (MoE) \cite{jacobs1991adaptive} inside the bottleneck of LoRA, VeRA~\cite{kopiczko2023elora} fixes the low-rank matrices and only tunes two vectors, MoLE~\cite{wu2023mole} directly uses multiple LoRA which combined by a gating function, SoRA~\cite{ding2023sparse} enables dynamic adjustments of the rank by using a gate unit, and DyLoRA~\cite{valipour2023dylora} follows the idea of nested dropout to train LoRA in a wide range of ranks.
Those methods achieve competitive performance and high parameter efficiency in single-task fine-tuning. However, those methods are not suitable for multi-task learning settings, since they do not consider shared information between multiple tasks. In contrast, the proposed TA-LoRA method can leverage task-shared information to enhance fine-tuning performance across various tasks.

\paragraph{Multi-Task Learning.}
As a widely used paradigm, Multi-Task Learning (MTL) aims to improve the average performance of a model by simultaneously learning multiple downstream tasks. 
To enhance the efficacy of learning multiple tasks simultaneously, some studies focus on decoupling task-shared and task-specific information through manual design~\cite{misra2016cross,liu2019end,invpt2022,taskprompter2023,gao2019nddr} or automatic architecture learning~\cite{guo2020learning,huang2018gnas,raychaudhuri2022controllable,sun2020adashare}. Other approaches propose balancing the losses or gradients of different tasks during training to avoid conflicts between them~\cite{chen2018gradnorm,yu2020gradient,liu2021towards,liu2021conflict,navon2022multi}. Additionally, some works employ task grouping techniques to select related tasks for joint model training~\cite{fifty2021efficiently,song2022efficient,standley2020tasks,zamir2018taskonomy}.
With the impressive generalization capability of large-scale pre-trained foundation models on downstream tasks, various multi-task parameter-efficient fine-tuning methods~\cite{liu2022polyhistor,liu2023hierarchical} have been proposed. For example, Polyhistor~\cite{liu2022polyhistor} designs a lightweight hyper-networks for hierarchical vision transformer, and HiPro~\cite{liu2023hierarchical} uses hierarchical prompt tuning to adapt pre-trained vision-language models.
Different from the previous works on multi-task learning, we leverage the powerful SAM and modify it for multi-task learning problems in computer vision. Based on mSAM, the proposed TA-LoRA method can effectively capture both the task-shared and task-specific information by fine-tuning update parameter tensors only.

\section{Methodology}
\begin{figure}[tb]
  \centering
\subfigure[The Original SAM.]
{\label{fig:single-sam}\includegraphics[width=0.9\textwidth]{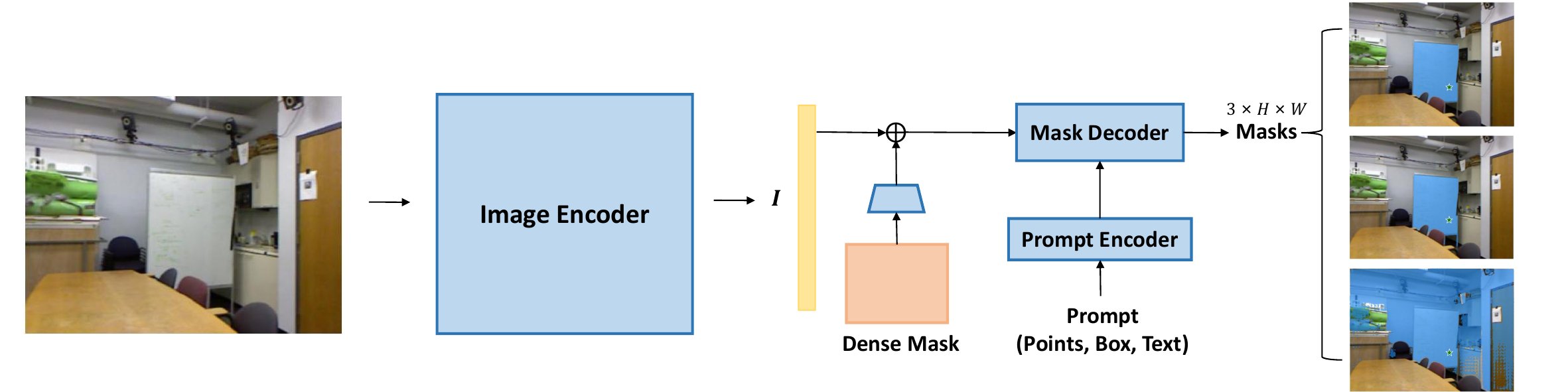}}
\subfigure[The mSAM with TA-LoRA for multiple downstream tasks.]
{\label{fig:multi-sam}\includegraphics[width=0.9\textwidth]{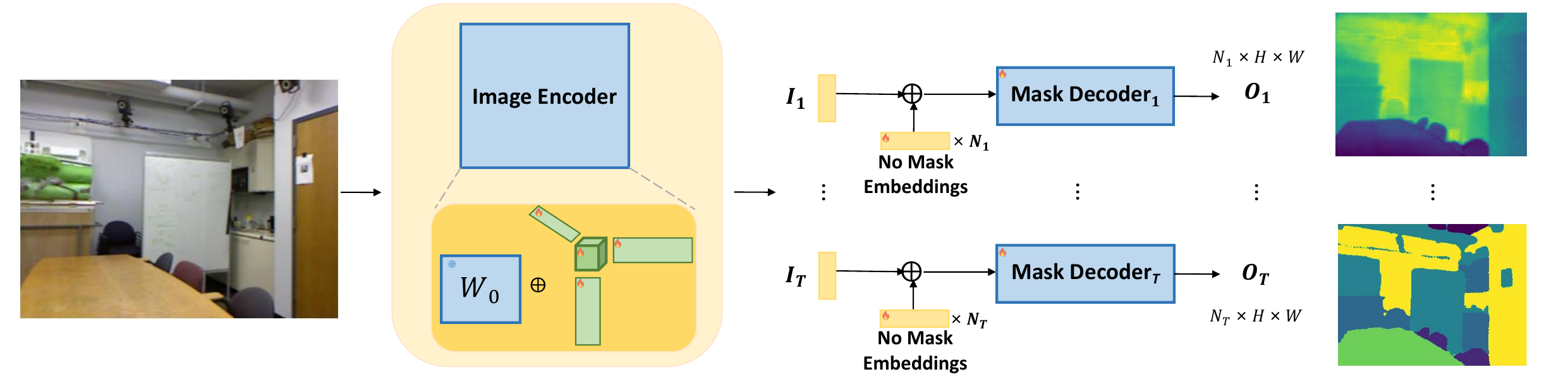}}
  \caption{An overview of the original SAM and the proposed mSAM with TA-LoRA. Compared to the original SAM, we froze the heavyweight image encoder while using TA-LoRA to fine-tune the update parameter tensors, and generate task-specific image embeddings for each task. Additionally, mSAM does not utilize the prompt encoder from the original SAM. Instead, we introduce trainable "no mask" embeddings with the corresponding number of output channels. Through these modifications, we can adapt to tasks with varying numbers of output channels. Moreover, mSAM has a task-specific mask decoder for each task.}
  \label{fig:overview}
  
    \vskip -0.1in
\end{figure}

In this section, we review SAM and LoRA in Section \ref{sec:pre}. 
Then, we introduce the proposed TA-LoRA method and how to employ the TA-LoRA method to mSAM in Sections \ref{sec:ta-lora} and \ref{sec:mod_sam}, respectively.
Figure \ref{fig:multi-sam} gives an illustration of the mSAM with TA-LoRA.

\subsection{Preliminaries}
\label{sec:pre}
\paragraph{SAM.} As illustrated in Figure \ref{fig:single-sam}, SAM consists of three main modules: a heavyweight image encoder, a prompt encoder, and a lightweight mask decoder. SAM first utilizes the image encoder to extract image features from images and the prompt encoder to encode different types of prompts (\ie, points, boxes, and masks) into prompt features. After that, the mask decoder will predict the final segmentation mask by performing attention-based feature interactions on image features and prompt features. By pre-training on the SA-1B dataset comprising over 11 million images and 1.1 billion masks generated automatically, SAM has demonstrated a remarkable zero-shot capability on the segmentation task. This capability reveals the potential of SAM as a foundation model for other tasks (\eg depth prediction, and surface normal).

\paragraph{LoRA.} The LoRA method~\cite{hu2022lora} assumes that each update parameter matrix have a low intrinsic rank and proposes to fine-tune them by freezing the pre-trained model. Formally, for a given task $t$, LoRA parameterizes an update parameter matrix $\Delta W_t \in \mathbb{R}^{d \times k}$ corresponding to a pre-trained parameter matrix $W_0 \in \mathbb{R}^{d \times k}$ by the product of two low-rank matrices, expressed as $B_tA_t$, where $B_t \in \mathbb{R}^{d \times r}$ and $A_t \in \mathbb{R}^{r \times k}$ with $r \ll \min(d, k)$. Thus, for a hidden input $x$, the output $h$ can be calculated by
\begin{equation}
    h=W'_tx=W_0x+\Delta W_tx=W_0x+B_tA_tx,
\end{equation}
where the $W'_t\in \mathbb{R}^{d \times k}$ denotes the parameter matrix after the update. 

LoRA has been proven as an efficient and effective approach in fine-tuning large pre-trained models for specific downstream tasks. Therefore, we consider LoRA as an important baseline method in our experiments. There are two different approaches to directly applying LoRA to the multi-task learning setting for SAM. One approach is using a hard parameter-sharing strategy, where all tasks use one shared LoRA matrix $\Delta W$. However, this hard parameter-sharing strategy may lead to imbalanced performance on all the tasks due to the competition among tasks for the shared LoRA parameters~\cite{zhang2021survey}. Another approach is to train a task-specific LoRA for each task and hence each task $t$ uses its own $\Delta W_t$. However, this approach can not harness the inter-task shared information necessary for fine-tuning across multiple tasks. 

\subsection{Task-Aware Low-Rank Adaptation}
\label{sec:ta-lora}
Suppose we are given $T$ tasks. For simplicity, we only consider the case that each task has only one layer to be fine-tuned in a PEFT way and it is easy to extend to multiple layers. 
For task $t$, its update parameter matrix is denoted by $\Delta W_t\in \mathbb{R}^{d \times k}$. 
Given the $T$ tasks, it is natural to aggregate all the update parameter matrices of all the tasks as an update parameter tensor $\Delta \mathbf{W}=\{\Delta W_1, \dots, \Delta W_T\} \in \mathbb{R}^{d \times k \times T}$. 

Inspired by LoRA, we place a low-rank assumption on $\Delta \mathbf{W}$. 
Such an assumption could hold due to the inter-task relatedness and intra-task low rank in each $\Delta W_t$ as LoRA did. Specifically, since different tasks in multi-task learning are usually assumed to be related, different update parameter matrices $\{\Delta W_t\}_{t=1}^T$ could be correlated, making $\Delta \mathbf{W}$ likely to be low-rank along the task axis (\ie the last axis). In this sense, this low-rank assumption on $\Delta \mathbf{W}$ could be viewed as a generalization of the low-rank assumption on the parameter matrix of linear models \cite{zhang2021survey}.

To achieve a low-rank $\Delta \mathbf{W}$, we parameterize it via tensor decomposition \cite{papalexakis2016tensors}, which is a technique to decompose a tensor into several low-rank factors. There are several tensor decomposition methods (\eg tensor-train decomposition, CP decomposition, and Tucker decomposition) and we choose the Tucker decomposition as it has a good representation ability.

\begin{figure}[tb]
  \centering
  \subfigure[LoRA for multiple tasks.]
{\label{fig:lora}\includegraphics[width=0.4\textwidth]{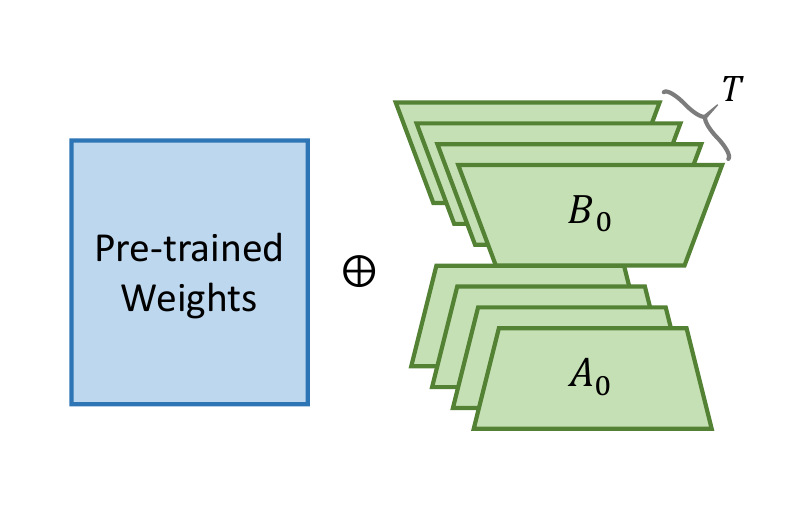}}
\subfigure[TA-LoRA for multiple tasks.]
{\label{fig:ta-lora}\includegraphics[width=0.4\textwidth]{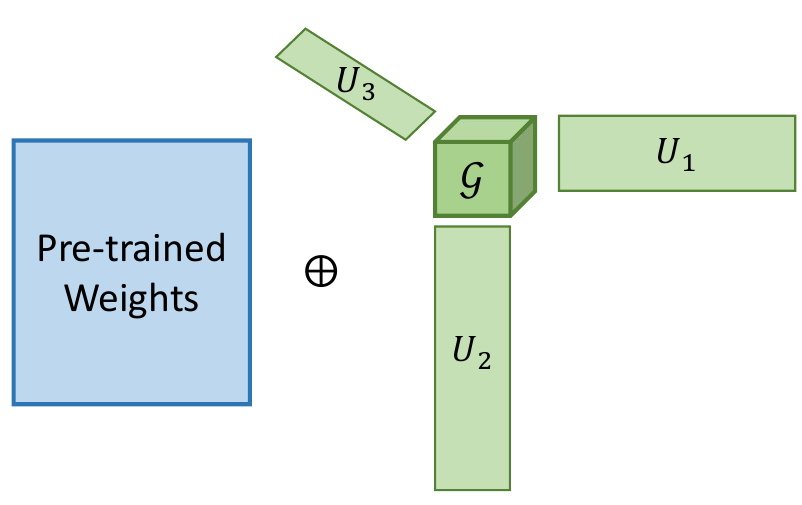}}
  \caption{Comparison between (a) LoRA and (b) TA-LoRA. Both LoRA and TA-LoRA employ low-rank approximation operations on the update parameter matrices. LoRA uses separate low-rank matrices to approximate each task's update parameter matrix, while TA-LoRA concatenates the update parameter matrices of each task into an update parameter tensor and applies low-rank tensor decomposition for approximation.}
  \label{fig:lora&tucker}
  
    \vskip -0.1in
\end{figure}

Specifically, we decompose the three-mode update parameter tensor $\Delta \mathbf{W} \in \mathbb{R}^{d \times k \times T}$ into a core tensor $\mathcal{G} \in \mathbb{R}^{p \times q \times v}$ and three factor matrices $U_1 \in \mathbb{R}^{d \times p}$, $U_2 \in \mathbb{R}^{k \times q}$ and $U_3 \in \mathbb{R}^{T \times v}$, where $p$, $q$, and $v$ denote the dimensions of factor matrices. Here we usually have $p, q, v \ll \min(d, k)$.
Formally, we have  
\begin{equation}
    \Delta \mathbf{W} = \mathcal{G} \times_1 U_1 \times_2 U_2 \times_3 U_3,
\label{eq:tucker}
\end{equation}
where $\times_n$ denotes the $n$-mode product. For $\mathcal{X} \in \mathbb{R}^{I_1 \times I_2 \times \dots \times I_m}$ and $Y \in \mathbb{R}^{I_n \times J}$, the $n$-mode product between them (\ie $\mathcal{Z}=\mathcal{X}\times_nY$) is defined as
\begin{equation}
    \mathcal{Z}(i_1, \dots, i_{n - 1}, j, i_{n + 1}, \dots, i_m) = \sum_{l=1}^{I_n}\mathcal{X}(i_1, \dots, i_{n - 1}, l, i_{n + 1}, \dots, i_m)Y(l, j),
\end{equation}
where $\mathcal{Z} \in \mathbb{R}^{I_1 \times \dots \times I_{n-1} \times J \times I_{n+1} \times \dots \times I_m}$.
Correspondingly, the $(i,j,t)$-th entry in $\Delta\mathbf{W}$ (\ie $\Delta\mathbf{W}(i, j, t)$) defined in Eq.~\ref{eq:tucker} can be written as
\begin{equation}
    \Delta\mathbf{W}(i, j, t) = \sum_{m=1}^{p}\sum_{n=1}^{q}\sum_{l=1}^{v}\mathcal{G}(m,n,l)U_1(i,m)U_2(j,n)U_3(t,l),
\end{equation}
where $i \in \{1,2, \dots, d\}$, $j \in \{1, 2, \dots, k\}$, and $t \in \{1, 2, \dots, T\}$ denote the indices of three mode, respectively. 
According to Eq.~\ref{eq:tucker}, factor matrices $U_1$ and $U_2$ keep task-shared information from different tasks, factor matrix $U_3$ keeps task-specific information, and $\mathcal{G}$ is a low-rank compression of the original update parameter tensor \cite{papalexakis2016tensors}. Hence, through the Tucker decomposition, the TA-LoRA method could capture both task-shared and task-specific information.

To reduce the redundancy in $U_1$, $U_2$ and $\mathcal{G}$, 
we utilize the orthogonal regularization to enforce the orthogonality of $U_1$, $U_2$ and core tensor $\mathcal{G}$ through the final dimension as
\begin{equation}
    R(U_1, U_2, \mathcal{G})=\|U_1^T U_1 - I\|_F^2 + \|U_2^T U_2 - I\|_F^2 + \sum_{l=1}^v\|\mathcal{G}(:, :, l)^T\mathcal{G}(:, :, l) - I\|_F^2,
\end{equation}
where $\|\cdot\|_F$ denotes the Frobenius norm for matrices. Therefore, the overall objective function of the TA-LoRA method is formulated as
\begin{equation}
    \mathcal{L}_{total} = \frac{1}{\sum w_i}\sum_{i=1}^Tw_i\mathcal{L}_i + \lambda R(U_1, U_2, \mathcal{G}),
\end{equation}
where $\mathcal{L}_i$ denotes the loss of task $i$, $w_i$ denotes the loss weight of task $i$, and $\lambda$ is the hyper-parameter that controls the impact of orthogonal regularization.
Specifically, for task $i$, $\mathcal{L}_i$ is formulated as
\begin{equation}
    \mathcal{L}_i = \frac{1}{n_i}\sum_{j=1}^{n_i} \ell_i(y_i^j, f(x_i^j)),
\end{equation}
where $x_i^j$ denotes the $j$-th training sample in task $i$, $y_i^j$ denotes the true output of $x_i^j$, $f(\cdot)$ denotes the mSAM as introduced in the next section, and $\ell_i$ denotes the loss function for task $i$.

For the standard LoRA method, the update parameter matrix $\Delta W$ is initialized as $\mathbf{0}$ by initializing $A$ from a Gaussian distribution and initializing $B$ as a zero matrix $\mathbf{0}$. Inspired by that, for the proposed TA-LoRA method, the core tensor $\mathcal{G}$ is initialized as $\mathbf{0}$, while factor matrices $U_1$, $U_2$, and $U_3$ are randomly initialized from the standard Gaussian distribution. Thus, for each task $t$, the update parameter matrix $\Delta W_t$ is $\mathbf{0}$ at the beginning of training.

During training, we utilize Eq. \ref{eq:tucker} to obtain the update parameter tensor $\mathbf{W}$ based on $U_1$, $U_2$, $U_3$, and $\mathcal{G}$ and employ $h=W'_tx=W_0x+\Delta \mathbf{W}(:,:,t)x$ on the forward process for task $t$. During the back-propagation process, we freeze the pre-trained matrix $W_0$ and only update $U_1$, $U_2$, $U_3$, and $\mathcal{G}$. During inference, we can store the updated parameter matrix of task $t$ as $W_t = W_0 + \Delta \mathbf{W}(:,:,t)x$. Thus, there is no additional latency introduced during inference.

\paragraph{Parameter Complexity.} We provide a parameter complexity comparison of the LoRA method and the proposed TA-LoRA method under the multi-task learning setting. To fine-tune the pre-trained matrix $W_0 \in \mathbb{R}^{d \times k}$, the LoRA method decomposes each update parameter matrix $\Delta W_t$ as $B_tA_t$, where $B_t \in \mathbb{R}^{d \times r}$ and $A_t \in \mathbb{R}^{r \times k}$. Therefore, for $T$ tasks, the parameter complexity of LoRA is $\mathcal{O}(Trd+Trk)$. For the proposed TA-LoRA method, we decompose the update parameter tensor $\Delta \mathbf{W}\in \mathbb{R}^{d \times k \times T}$ as $\mathcal{G} \times_1 U_1 \times_2 U_2 \times_3 U_3$. Therefore, the parameter complexity of the proposed method is $pqv + dp + kq + Tv \sim \mathcal{O}(dp + kq)$ since $T, p, q, v \ll \min(d, k)$. This implies that the parameter complexity of LoRA increases linearly with the number of tasks $T$, while the proposed TA-LoRA method exhibits a sublinear complexity, thereby demonstrating the parameter efficiency of the proposed  TA-LoRA method.

\subsection{mSAM} \label{sec:mod_sam}
Despite the tremendous potential exhibited by SAM as a fundamental visual model, its reliance on prompt-guided mask generation presents challenges in achieving end-to-end adaptability to downstream tasks with varying numbers of output channels.

To accommodate downstream tasks with varying numbers of output channels, as depicted in Figure \ref{fig:multi-sam}, the proposed mSAM introduces trainable no mask embeddings with distinct numbers of output channels and consequently removes dense mask embeddings and the prompt encoder from SAM. Formally, we add the image embedding $I_t \in \mathbb{R}^{C \times H \times W}$ of task $t$ to no mask embeddings $E_t \in \mathbb{R}^{N_t \times C \times H \times W}$ and the result will be $(I_t + E_t) \in \mathbb{R} ^ {N_t \times C \times H \times W}$, which can be fed into mask decoder of task $t$ and generate the final prediction $O_t \in \mathbb{R}^{N_t \times H \times W}$.
Furthermore, we froze the pre-trained heavyweight image encoder and use the TA-LoRA method to fine-tune the update parameter tensors in the self-attention module (\ie the query, key, and value) 
placed in each layer of the image encoder.
Moreover, inspired by \cite{yuan2024fulllora}, we perform fine-tuning on the scale and bias parameters within the layer normalization layers of the image encoder and this adjustment introduces a minimal number of learnable parameters while significantly enhancing the performance. 
Additionally, we duplicate the mask decoder as a task-specific mask decoder for each task and fully fine-tune them as each of them is a lightweight module.

\section{Experiments}

In this section, we empirically evaluate the proposed TA-LoRA for SAM on two multi-task learning benchmark datasets, including \textbf{NYUv2} \cite{silberman2012indoor} and \textbf{CityScapes} \cite{Cordts2016Cityscapes}, with various tasks, \eg semantic segmentation, depth prediction, surface normal estimation tasks.

\paragraph{Baselines.}
The proposed method is evaluated against several baselines that have a similar number of trainable parameters, including Single-Task Learning (STL) and three conventional multi-task learning architectures with a similar number of trainable parameters: Hard-Parameter Sharing (HPS) which trains a task-shared encoder and multiple task-specific decoders, Cross-Stitch Network~\cite{misra2016cross} which uses cross-stitch unit to combine the activations from multiple networks, Multi-Task Attention Network (MTAN)~\cite{liu2019end} which uses task-specific soft-attention modules to capture task-specific features from the global pool and NDDR-CNN~\cite{gao2019nddr} which combines existing CNN components by neural discriminative dimensionality reduction.
Additionally, we compare the proposed method with the LoRA-STL which trains a task-specific LoRA for each task, and LoRA-HPS which shares only one LoRA for all tasks.

For traditional MTL methods (\ie STL, HPS, Cross-Stitch Network, MTAN, and NDDR-CNN), we use the Deeplab-ResNet~\cite{chen2018encoder} with atrous convolutions, a popular architecture for pixel-wise prediction tasks, as encoders and the Atrous Spatial Pyramid Pooling (ASPP) architecture~\cite{chen2018encoder} as decoders. We adopt the pre-trained ResNet-50 to implement the Deeplab-ResNet~\cite{chen2018encoder}.
For the methods of fine-tuning mSAM (\ie LoRA-STL, LoRA-HPS, and TA-LoRA), we use SAM-L which adopts ViT-L~\cite{dosovitskiy2020image} as the image encoder.

\paragraph{Evaluation metric.}
For the \textbf{NYUv2} and \textbf{CityScapes} datasets which have multiple metrics for each task, following the setup of \cite{maninis2019attentive}, we use the average of the relative improvement of each task over the HPS architecture as the evaluation metric, which is formulated as
$$\Delta_{b}=\frac{1}{T}\sum\limits_{i=1}^{T}\frac{1}{N_{i }}\sum\limits_{j=1}^{N_{i}}\frac{(-1)^{s_{i, j}}(M_{i, j}^{b}-M_{i, j}^{HPS})}{M_{i, j}^{HPS}},$$
where $T$ denotes the number of tasks, $N_i$ denotes the number of metrics for task $i$, $M_{i,j}^{b}$ and $M_{i, j}^{HPS}$ denote the performance of the method $b$ and the HPS architecture for the $j$th metric in task $i$ respectively. $s_{i, j}$ is set to 1 if a lower value indicates better performance for the $j$th metric in task $i$ and otherwise 0.

Specifically, for the semantic segmentation task, we use the mean Intersection over Union (mIoU) and Pixel Accuracy (Pix Acc) to evaluate. For the depth prediction task, we use the Absolute Error (Abs Err) and Real Error (Rel Err) to evaluate. For the surface normal estimation task, we use the mean and the median of angular error measured in degrees and the percentage of pixels whose angular error is within $11.25$, $22.5$, and $30$ degrees to evaluate.

\subsection{NYUv2}\label{sec:nyuv2}
The \textbf{NYUv2} dataset \cite{silberman2012indoor} consists of video sequences of various indoor scenes recorded by RGB and Depth cameras in Microsoft Kinect. It contains 1,449 images with ground truth, where 795 images are for training and 654 images are for validation. This dataset has three tasks: 13-class semantic segmentation, depth estimation, and surface normal prediction.

\paragraph{Setups.}
The batch size is set to $4$. The cross-entropy loss, $L_1$ loss, and cosine similarity loss are used as the loss functions of the semantic segmentation, depth estimation, and surface normal prediction tasks, respectively. The Adam optimizer is used for updating fine-tuned parameters. In the Adam optimizer, an initial learning rate is set to $10^{-3}$, the linear learning rate scheduler with warmup is adopted while the warmup rate is set to $0.05$, and the weight decay is set to $10^{-6}$. The dropout rate of LoRA-STL, LoRA-HPS, and TA-LoRA is set to $0.1$. For LoRA-STL, we try two different ranks as $r=16$ and $r=32$ while for LoRA-HPS, we set rank $r$ to $32$. For the proposed TA-LoRA, we set $p=q=32$, $v=8$. The hyperparameter $\lambda$ is set to $1$. The total number of fine-tuned epochs is set to 200.

\begin{table*}[tb]
        \centering
        \caption{Performance on three tasks (\ie 13-class semantic segmentation, depth estimation, and surface normal prediction) in the \textbf{NYUv2} dataset. The best results for each task are shown in \textbf{bold}.$\uparrow$($\downarrow$) means that the higher (lower) the value, the better the performance. The number of trainable parameters (\ie, Params.) is calculated in MB.}
    	\resizebox{\linewidth}{!}{
            \begin{tabular}{cccccccccccc}
                \toprule
                \multirow{4}{*}{\textbf{Method}} & \multicolumn{2}{c}{\textbf{Segmentation}} & \multicolumn{2}{c}{\textbf{Depth}} & \multicolumn{5}{c}{\textbf{Surface Normal}} & \multirow{4}{*}{\textbf{Param. (M)$\downarrow$}} & \multirow{4}{*}{\textbf{$\Delta \uparrow$}}\\
    			\cmidrule(r){2-3} \cmidrule(r){4-5} \cmidrule(r){6-10} &\multirow{2.5}{*}{\textbf{mIoU${\uparrow}$}} &  \multirow{2.5}{*}{\textbf{Pix Acc $\uparrow$}} &  \multirow{2.5}{*}{\textbf{Abs Err $\downarrow$}} &  \multirow{2.5}{*}{\textbf{Rel Err$\downarrow$}} & \multicolumn{2}{c}{\textbf{Angle Distance}} & \multicolumn{3}{c}{\textbf{Within $t^{\circ}$}} \\ \cmidrule(r){6-7} \cmidrule(r){8-10} & & & & & \textbf{Mean $\downarrow$} & \textbf{Median $\downarrow$}  & \textbf{11.25 $\uparrow$} & \textbf{22.5 $\uparrow$} & \textbf{30 $\uparrow$}  \\
    			\midrule
    			HPS & 54.48 & 75.82 & 0.3839 & 0.1548 & 23.50 & 17.06 & 35.31 & 61.10 & 72.14 & 71.89 & $+0.00\%$ \\
                STL & 53.98 & 75.38 & 0.3945 & 0.1631 & 22.25 & 15.63 & 38.12 & 64.38 & 74.81 & 118.91 & $+0.45\%$\\
                Cross-Stitch & 53.46 & 75.49 & 0.3804 & 0.1555 & 23.01 & 16.33 & 37.01 & 62.42 & 73.02 & 118.89 & $+0.66\%$\\
                MTAN & 54.74 & 75.78 & 0.3796 & 0.1549 & 22.97 & 16.30 & 36.91 & 62.63 & 73.32 & 92.35 & $+0.77\%$\\
                NDDR-CNN & 53.84 & 75.23 & 0.3871 & 0.1560 & 22.60 & 16.07 & 37.67 & 63.43 & 73.92 & 169.10 & $+0.91\%$\\
                \midrule
                LoRA-HPS (r=32) & 56.77 & 78.37 & 0.3470 & 0.1412 & 18.97 & 13.41 & 44.56 & 71.20 & 80.68 & 58.84 & $+10.67\%$\\
                LoRA-STL (r=16) & 62.06 & 81.72 & 0.3124 & 0.1233 & 16.44 &  11.39 & 51.04 & 77.01 & 85.31 & 64.83 & $+20.25\%$\\
                LoRA-STL (r=32) & 58.34 & 78.61 & 0.3330 & 0.1335 & 16.54 & 11.42 & 51.08 & 76.65 & 84.99 & 82.84 & $+16.34\%$\\
                \midrule
                TA-LoRA & \textbf{65.98} & \textbf{83.42} & \textbf{0.2898} & \textbf{0.1140} & \textbf{16.34} & \textbf{11.33} & \textbf{51.22} & \textbf{77.20} & \textbf{85.51} & 59.59 & $\mathbf{+23.93\%}$\\
                \bottomrule
            \end{tabular}
            }
        \label{tab:nyuv2}
        
    \end{table*}

\begin{figure}[!htb]
  \centering
  \includegraphics[width=0.95\linewidth]{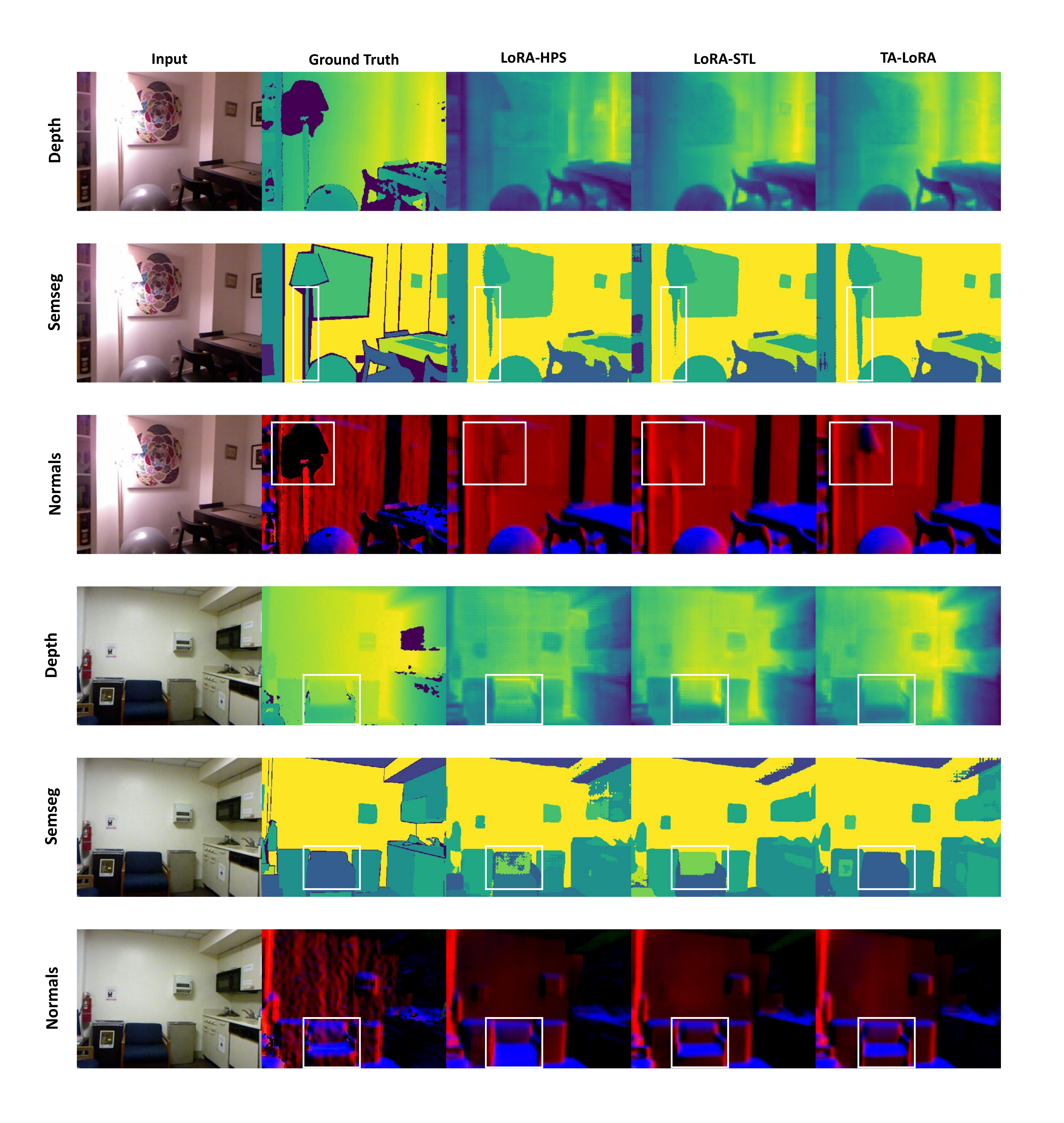}
  \vskip -0.1in
  \caption{
  Generation quality of mSAM fine-tuned by LoRA-HPS, LoRA-STL, and TA-LoRA on \textbf{NYUv2} dataset. It can be observed that our method can generate outputs of higher quality, particularly in the regions in the white bounding box.
  }
  \label{fig:quality}
  
    \vskip -0.12in
\end{figure}

\paragraph{Results.}
The results on the \textbf{NYUv2} dataset are shown in Table \ref{tab:nyuv2}. As can be seen, we can observe that fine-tuning mSAM using our proposed TA-LoRA method not only achieves optimal average performance but also attains the best results on each task.
Additionally, compared to training with a conventional multi-task learning architecture that utilizes convolutional neural networks, fine-tuning mSAM leads to significant improvements, highlighting the potential of the mSAM in multi-task learning.
Furthermore, we observed that training a specific LoRA for each task (\ie LoRA-STL) yields better results compared to training a shared LoRA (\ie LoRA-HPS) which demonstrates the importance of task-specific information during training a large foundation model. The superior performance of TA-LoRA over LoRA-STL and LoRA-HPS suggests that TA-LoRA can leverage both task-shared and task-specific information well, thereby improving the model's performance.
Figure \ref{fig:quality} shows the output predictions of mSAM fine-tuned with LoRA-STL, LoRA-HPS, and TA-LoRA, respectively. As can be seen, fine-tuning mSAM with TA-LoRA has better visualization compared to the baselines. For example, in the white boxes, we observe that our proposed method generates more accurate results compared to the baseline methods when dealing with ambiguous objects.

\subsection{CityScapes}
The \textbf{CityScapes} dataset \cite{Cordts2016Cityscapes} is a comprehensive urban street scene understanding dataset and consists of a diverse set of stereo video sequences captured from 50 distinct cities under favorable weather conditions during daylight hours. The dataset comprises 2,975 annotated images for training and an additional 500 images for testing. Following the setup of \cite{liu2019end}, we conduct experiments on two tasks: 7-class semantic segmentation and depth estimation.

\paragraph{Setups.}
The batch size is set to $8$. The cross-entropy loss and $L_1$ loss are used as the loss function of the semantic segmentation and depth estimation tasks, respectively. The Adam optimizer is used for updating fine-tuned parameters. In the Adam optimizer, an initial learning rate is set to $10^{-3}$, the linear learning rate scheduler with warmup is adopted while the warmup rate is set to $0.05$, and the weight decay is set to $10^{-6}$. The dropout rate of LoRA-STL, LoRA-HPS, and TA-LoRA is set to $0.1$. For LoRA-STL, we try two different ranks as $r=8$ and $r=16$ while for LoRA-HPS, we set rank $r$ to $16$. For the proposed TA-LoRA, we set $p=q=16$, $v=4$. The hyperparameter $\lambda$ is set to $1$. The total number of fine-tuned epochs is set to 50.

\paragraph{Results.}

\begin{table*}[!t]
        \centering
        \caption{Performance on two tasks (\ie 7-class semantic segmentation and depth estimation) in the \textbf{CityScapes} dataset. The best results for each task are shown in \textbf{bold}.$\uparrow$($\downarrow$) means that the higher (lower) the value, the better the performance. The number of trainable parameters (\ie, Params.) is calculated in MB.}
    	\resizebox{0.75\linewidth}{!}{
            \begin{tabular}{ccccccc}
                \toprule
                \multirow{2.5}{*}{\textbf{Method}} & \multicolumn{2}{c}{\textbf{Segmentation}} & \multicolumn{2}{c}{\textbf{Depth}} & \multirow{2.5}{*}{\textbf{Param. (M)$\downarrow$}} & \multirow{2.5}{*}{\textbf{$\Delta \uparrow$}}\\
    			\cmidrule(r){2-3} \cmidrule(r){4-5} &{\textbf{mIoU${\uparrow}$}} &  {\textbf{Pix Acc $\uparrow$}} &  {\textbf{Abs Err $\downarrow$}} &  {\textbf{Rel Err$\downarrow$}}\\
    			\midrule
    			HPS & 67.40 & 90.92 & 0.0142 & 45.4262 & 55.76 & $+0.00\%$\\
                STL & 68.13 & 91.28 & 0.0133 & 45.0390 & 79.27 & $+2.17\%$\\
                Cross-Stitch & 68.01 & 91.29 & 0.0135 & 44.4246 & 79.27 & $+2.11\%$\\
                MTAN & 68.97 & 91.59 & 0.0136 & 43.7508 & 72.04 & $+2.74\%$\\
                NDDR-CNN & 68.02 & 91.25 & 0.0137 & 44.8662 & 101.58 & $+1.51\%$ \\
                \midrule
                LoRA-HPS (r=16) & 86.23 & 96.30 & 0.0123 & 34.2000 & 37.35 & $+17.98\%$\\
                LoRA-STL (r=8) & 85.86 & 96.26 & 0.0110 & 34.3000 & 37.35 & $+20.07\%$\\
                LoRA-STL (r=16) & 82.64 & 95.28 & \textbf{0.0107} & 33.4312  & 43.35 & $+19.62\%$\\
                \midrule
                TA-LoRA & \textbf{87.45} & \textbf{96.80} & 0.0113 & \textbf{33.0086} & 37.44 & $\mathbf{+20.99\%}$ \\
                \bottomrule
            \end{tabular}
            }
        \label{tab:cityscape}
    \end{table*}

The results on the \textbf{CityScapes} dataset are shown in Table \ref{tab:cityscape}. As can be seen, we can observe that fine-tuning mSAM using our proposed TA-LoRA method not only achieves optimal average performance but also outperforms other methods in most of the evaluated metrics for each task, especially in the metrics of semantic segmentation task.
Furthermore, the significant improvements achieved by fine-tuning mSAM compared to conventional multi-task learning architectures underscore the potential of mSAM in the context of multi-task learning.
Moreover, we observed that training task-specific LoRAs (\ie LoRA-STL) yields better results compared to training a shared LoRA (\ie LoRA-HPS).
This observation implies the necessity of task-specific information.
The superior performance of TA-LoRA over LoRA-STL and LoRA-HPS suggests that TA-LoRA can utilize both task-shared and task-specific information well through tensor decomposition, leading to improvement of the performance.

\subsection{PASCAL-Context}
We conduct experiments on the \textbf{PASCAL-Context} dataset~\cite{everingham2010pascal}. 
The \textbf{PASCAL-Context} dataset comprises 4,998 images for training and 5,105 images for testing. This dataset offers dense labels for four tasks including semantic segmentation, human parsing, surface normal estimation, and saliency detection while the pseudo ground truth labels of surface normal estimation and saliency detection are provided by \cite{maninis2019attentive}. Following the setup of \cite{liu2022polyhistor}, we use the mean intersection-over-union (mIoU) to evaluate the performance on semantic segmentation, human part segmentation, and saliency detection. Additionally, we use the mean error (mErr) to evaluate the performance on surface normal estimation.

\paragraph{Baselines.}
The proposed method is evaluated against several baselines that have a similar number of trainable parameters, including Single-Task Learning (STL) and four conventional multi-task learning architectures with a similar number of trainable parameters: Hard-Parameter Sharing (HPS) which trains a task-shared encoder and multiple task-specific decoders, Cross-Stitch Network~\cite{misra2016cross} which uses cross-stitch unit to combine the activations from multiple networks, Multi-Task Attention Network (MTAN)~\cite{liu2019end} which uses task-specific soft-attention modules to capture task-specific features from the global pool, and NDDR-CNN~\cite{gao2019nddr} which combines existing CNN components by neural discriminative dimensionality reduction. Furthermore, the proposed method is compared with three 
parameter-efficient fine-tuning methods that are designed for multi-task learning: Hyperformer~\cite{mahabadi2021parameter} which uses hypernetworks to generate the adapter parameters for all layers, Polyhistor~\cite{liu2022polyhistor} which only uses decomposed hypernetworks to generate the adapter parameters, and Polyhistor-Lite~\cite{liu2022polyhistor} which combines decomposed hypernetworks and layer-wise scaling kernels to generate the adapter parameters.

For traditional MTL methods (\ie STL, HPS, Cross-Stitch Network, MTAN, and NDDR-CNN), we use the Deeplab-ResNet~\cite{chen2018encoder} with atrous convolutions, a popular architecture for pixel-wise prediction tasks, as encoders and the Atrous Spatial Pyramid Pooling (ASPP) architecture~\cite{chen2018encoder} as decoders. We adopt the pre-trained ResNet-50 to implement the Deeplab-ResNet~\cite{chen2018encoder}.
For Hyperformer and Polyhistor, we use Swin Transformer~\cite{liu2021Swin} as the encoder and All-MLP decoder of Segformer~\cite{xie2021segformer} as the decoder.
For mSAM with TA-LoRA, we use SAM-L which adopts ViT-L~\cite{dosovitskiy2020image} as the image encoder.

\paragraph{Setups.}
The batch size is set to $8$. The cross-entropy loss, $L_1$ loss, and cosine similarity loss are used as the loss functions of the semantic segmentation, depth estimation, and surface normal prediction tasks, respectively. The Adam optimizer is used for updating fine-tuned parameters. In the Adam optimizer, an initial learning rate is set to $10^{-3}$, the linear learning rate scheduler with warmup is adopted where the warmup rate is set to $0.05$, and the weight decay is set to $10^{-6}$. The dropout rate of TA-LoRA is set to $0.1$. For the proposed TA-LoRA, we set $p$ and $q$ to be 16 and  $v$ to be 4. The hyperparameter $\lambda$ is set to $1$ by default. The total number of fine-tuning epochs is set to 30.

\begin{table*}[!htb]
        \centering
        \caption{Performance on four tasks (\ie 21-class semantic segmentation, 7-class human parts segmentation, saliency estimation, and surface normal estimation) in the \textbf{PASCAL-Context} dataset. The best results for each task are shown in \textbf{bold}. $\uparrow$($\downarrow$) means that the higher (lower) the value, the better the performance. The number of trainable parameters (\ie, Params.) is calculated in MB.}
    	\resizebox{0.6\linewidth}{!}{
            \begin{tabular}{ccccccc}
                \toprule
                \textbf{Method} & \textbf{Seg.$\uparrow$} & \textbf{H.Parts$\uparrow$} & \textbf{Sal.$\uparrow$} & \textbf{Normal$\downarrow$} & \textbf{Param. (M)} & $\Delta\uparrow$\\
                \midrule
                HPS &  64.77 & 57.91 & 64.10 & \textbf{14.21} & 30.07 & $+0.00\%$\\
                STL & 65.14 & 58.58 & 65.02 & 15.94 & 63.60 & $-2.25\%$\\
                Cross-Stitch & 64.97 & 58.63 & 64.46 & 15.32 & 79.46 & $-1.42\%$\\
                MTAN & 64.56 & 59.08 & 64.57 & 14.74 & 36.61 & $-0.33\%$\\
                NDDR-CNN & 65.28 & 59.18 & 65.09 & 15.57 & 69.25 & $-1.26\%$\\
                \midrule
                Hyperformer & 71.43 & 60.73 & 65.54 & 17.77 & 287.32 & $-1.91\%$\\
                Polyhistor & 70.87 & 59.54 & 65.47 & 17.47 & 34.18 & $-2.14\%$\\
                Polyhistor-Lite & 70.24 & 59.12 & 64.75 & 17.40 & 11.29 & $-2.72\%$\\
                \midrule
                TA-LoRA & \textbf{73.83} & \textbf{70.39} & \textbf{75.68} & 17.39 & 68.43 & $+7.80\%$\\
                \bottomrule
            \end{tabular}
            }
        \label{tab:pascal}
    \end{table*}

\paragraph{Results.}
The results on the \textbf{PASCAL-Context} dataset are shown in Table \ref{tab:pascal}. As can be seen, we can observe that fine-tuning mSAM using the proposed TA-LoRA method achieves the best average performance.
Additionally, compared with conventional multi-task learning architectures that utilizes convolutional neural networks and multi-task parameter-efficient fine-tuning method on Swin Transformer, fine-tuning mSAM leads to significant improvements, highlighting the potential of the mSAM in multi-task learning.

\subsection{Ablation Study}

\paragraph{Sensitive to rank.}

\begin{table*}[!htb]
        \centering
        \caption{Ablation studies on the impact of rank in the \textbf{NYUv2} dataset. The best results for each task are shown in \textbf{bold}. $\uparrow$($\downarrow$) means that the higher (lower) the value, the better the performance. The number of trainable parameters (\ie, Params.) is calculated in MB.}
    	\resizebox{\linewidth}{!}{
            \begin{tabular}{cccccccccccc}
                \toprule
                \multirow{4}{*}{\textbf{Method}} & \multicolumn{2}{c}{\textbf{Segmentation}} & \multicolumn{2}{c}{\textbf{Depth}} & \multicolumn{5}{c}{\textbf{Surface Normal}} & \multirow{4}{*}{\textbf{Param. (M)$\downarrow$}} & \multirow{4}{*}{\textbf{$\Delta \uparrow$}}\\
    			\cmidrule(r){2-3} \cmidrule(r){4-5} \cmidrule(r){6-10} &\multirow{2.5}{*}{\textbf{mIoU${\uparrow}$}} &  \multirow{2.5}{*}{\textbf{Pix Acc $\uparrow$}} &  \multirow{2.5}{*}{\textbf{Abs Err $\downarrow$}} &  \multirow{2.5}{*}{\textbf{Rel Err$\downarrow$}} & \multicolumn{2}{c}{\textbf{Angle Distance}} & \multicolumn{3}{c}{\textbf{Within $t^{\circ}$}} \\ \cmidrule(r){6-7} \cmidrule(r){8-10} & & & & & \textbf{Mean $\downarrow$} & \textbf{Median $\downarrow$}  & \textbf{11.25 $\uparrow$} & \textbf{22.5 $\uparrow$} & \textbf{30 $\uparrow$}  \\
    			\midrule
    			HPS & 54.48 & 75.82 & 0.3839 & 0.1548 & 23.50 & 17.06 & 35.31 & 61.10 & 72.14 & 71.89 & $+0.00\%$ \\
    			STL & 53.98 & 75.38 & 0.3945 & 0.1631 & 22.25 & 15.63 & 38.12 & 64.38 & 74.81 & 118.91 & $+0.45\%$\\
                \midrule
                 LoRA-HPS (r=32) & 56.77 & 78.37 & 0.3470 & 0.1412 & 18.97 & 13.41 & 44.56 & 71.20 & 80.68 & 58.84 & $+10.67\%$\\
                LoRA-STL (r=16) & 62.06 & 81.72 & 0.3124 & 0.1233 & 16.44 &  11.39 & 51.04 & 77.01 & 85.31 & 64.83 & $+20.25\%$\\
                 LoRA-STL (r=32) & 58.34 & 78.61 & 0.3330 & 0.1335 & 16.54 & 11.42 & 51.08 & 76.65 & 84.99 & 82.84 & $+16.34\%$\\
                \midrule
                TA-LoRA ($p=q=16, v=8$) & 64.66 & 83.15 & 0.2966 & 0.1153 & 16.40 & 11.42 & 50.89 & 76.85 & 85.25 & 53.03 & $+22.85\%$\\
                TA-LoRA ($p=q=32, v=4$) & 65.57 & 83.29 & \textbf{0.2888} & 0.1149 & \textbf{16.29} & \textbf{11.3}2 & \textbf{51.22} & \textbf{77.40} & \textbf{85.64} & 59.21 & $+23.77\%$\\
                TA-LoRA ($p=q=32, v=8$) & \textbf{65.98} & \textbf{83.42} & 0.2898 & \textbf{0.1140} & 16.34 & 11.33 & \textbf{51.22} & 77.20 & 85.51 & 59.59 & $+23.93\%$\\
                \bottomrule
            \end{tabular}
            }
        \label{tab:rank}
    \end{table*}
    
We explore the sensitivity of TA-LoRA to different combinations of ranks while keeping the remaining hyperparameters consistent with the previous settings as described in Section \ref{sec:nyuv2}. The results are shown in Table \ref{tab:rank}. As can be seen, our method consistently outperforms the LoRA-STL method and the LoRA-HPS across different combinations of ranks, demonstrating our approach's effectiveness.

\paragraph{Impact of orthogonal regularization.} We conducted an ablation experiment on \textbf{NYUv2} dataset to evaluate the impact of applying orthogonal constraints to the core tensor $\mathcal{G}$. We employed the same hyperparameter settings as described in Section \ref{sec:nyuv2}. The results are presented in Table \ref{tab:ablation_orthorgal}, demonstrating that decomposing the core tensor $\mathcal{G}$ into matrices along the task dimension and employing an orthogonality regularization term effectively improves the performance of the model across various tasks. Furthermore, we observed that our method consistently outperforms the LoRA-STL method across different hyperparameter settings, which shows the effectiveness of our approach.

\begin{table*}[!htb]
        \centering
        \caption{Ablation studies on the impact of orthogonal regularization in the \textbf{NYUv2} dataset. The best results for each task are shown in \textbf{bold}. $\uparrow$($\downarrow$) means that the higher (lower) the value, the better the performance}
    	\resizebox{\linewidth}{!}{
            \begin{tabular}{ccccccccccc}
                \toprule
                \multirow{4}{*}{\textbf{Method}} & \multicolumn{2}{c}{\textbf{Segmentation}} & \multicolumn{2}{c}{\textbf{Depth}} & \multicolumn{5}{c}{\textbf{Surface Normal}} & \multirow{4}{*}{\textbf{$\Delta \uparrow$}}\\
    			\cmidrule(r){2-3} \cmidrule(r){4-5} \cmidrule(r){6-10} &\multirow{2.5}{*}{\textbf{mIoU${\uparrow}$}} &  \multirow{2.5}{*}{\textbf{Pix Acc $\uparrow$}} &  \multirow{2.5}{*}{\textbf{Abs Err $\downarrow$}} &  \multirow{2.5}{*}{\textbf{Rel Err$\downarrow$}} & \multicolumn{2}{c}{\textbf{Angle Distance}} & \multicolumn{3}{c}{\textbf{Within $t^{\circ}$}} \\ \cmidrule(r){6-7} \cmidrule(r){8-10} & & & & & \textbf{Mean $\downarrow$} & \textbf{Median $\downarrow$}  & \textbf{11.25 $\uparrow$} & \textbf{22.5 $\uparrow$} & \textbf{30 $\uparrow$}  \\
    			\midrule
    			HPS & 54.48 & 75.82 & 0.3839 & 0.1548 & 23.50 & 17.06 & 35.31 & 61.10 & 72.14 & $+0.00\%$ \\
                STL & 53.98 & 75.38 & 0.3945 & 0.1631 & 22.25 & 15.63 & 38.12 & 64.38 & 74.81 & $+0.45\%$\\
                \midrule
                LoRA-HPS (r=32) & 56.77 & 78.37 & 0.3470 & 0.1412 & 18.97 & 13.41 & 44.56 & 71.20 & 80.68 & $+10.67\%$\\
                LoRA-STL (r=16) & 62.06 & 81.72 & 0.3124 & 0.1233 & 16.44 &  11.39 & 51.04 & 77.01 & 85.31 & $+20.25\%$\\
                LoRA-STL (r=32) & 58.34 & 78.61 & 0.3330 & 0.1335 & 16.54 & 11.42 & 51.08 & 76.65 & 84.99 & $+16.34\%$\\
                \midrule
                TA-LoRA ($U_1$ and $U_2$) & 65.22 & 83.11 & 0.2984 & 0.1195 & 16.56 & 11.53 & 50.59 & 76.63 & 85.03 & $+22.30\%$\\
                TA-LoRA ($U_1$, $U_2$ and $\mathcal{G}$) & \textbf{65.98} & \textbf{83.42} & \textbf{0.2898} & \textbf{0.1140} & \textbf{16.34} & \textbf{11.33} & \textbf{51.22} & \textbf{77.20} & \textbf{85.51} & $\mathbf{+23.93\%}$\\
                \bottomrule
            \end{tabular}
            }
        \label{tab:ablation_orthorgal}
        
    \vskip -0.1in
    \end{table*}

\section{Conclusion}
In this paper, we propose the Task-Aware Low-Rank Adaptation (TA-LoRA) method to fine-tune the modified Segment Anything Model (mSAM) for multi-task learning, which leverages a low-rank tensor decomposition method to use both task-shared and task-specific information.
Furthermore, we introduce no mask embeddings to guide mSAM in generating outputs with the corresponding channel number, enabling mSAM can be adapted to different tasks. To further improve the performance, we fine-tune the scale and bias parameters within the layer normalization layers of the image encoder and train a task-specific decoder for each task.
The experimental results demonstrate the effectiveness of fine-tuning mSAM with the TA-LoRA method and reveal the potential of mSAM for multi-task learning.

%
\bibliographystyle{splncs04}
\bibliography{main}



\end{document}